\documentclass[11pt,a4paper]{article}

\usepackage{emnlp2021}
\usepackage{times}
\usepackage{latexsym}

\usepackage{microtype}

\newcommand{\tablesize}{\fontsize{8.5pt}{10pt}\selectfont}

\usepackage[noabbrev,capitalize]{cleveref}
\usepackage{multirow}
\usepackage{graphicx}
\usepackage[normalem]{ulem}

\usepackage{comment}
\usepackage{adjustbox}
\usepackage{enumitem}

\usepackage{subfig}
\def\hideXXX#1{}
\def\XXXifaccepted#1{} % hide for now

\def\hidePV#1{}

\def\footurl#1{\footnote{\url{#1}}}

\def\parcite#1{\citep{#1}} % (Smith, 2012)
\def\inparcite#1{\citealp{#1}} % should be Smith, 2012
\def\furl#1{\footnote{\url{#1}}}

\title{Continuous Rating as Reliable Human Evaluation \\ of Simultaneous Speech Translation}

\author{D\'avid Javorsk\'y \\\And
  Dominik Mach\'a\v{c}ek \\
  \\
  Charles University, Faculty of Mathematics and Physics \\
  Institute of Formal and Applied Linguistics \\
  \texttt{\{surname\}@ufal.mff.cuni.cz} \\\And
  Ond\v{r}ej Bojar}

\date{}

\begin{document}
\maketitle
\begin{abstract}
Simultaneous speech translation (SST) can be evaluated on simulated online events where human evaluators watch subtitled videos and continuously express their satisfaction by pressing buttons (so called Continuous Rating). Continuous Rating is easy to collect, but little is known about its reliability, or relation to comprehension of foreign language document by SST users. In this paper, we contrast Continuous Rating with factual questionnaires on judges with different levels of source language knowledge. Our results show that Continuous Rating is easy and reliable SST quality assessment if the judges have at least limited knowledge of the source language. Our study indicates users' preferences on subtitle layout and presentation style and, most importantly, provides a significant evidence that users with advanced source language knowledge prefer low latency over fewer re-translations.
\end{abstract}

\section{Introduction}

Simultaneous speech translation (SST) is a technology that assists users to understand
and follow a speech in a foreign language in real-time. The users may need such
an assistance because of limited knowledge of the source language, the speaker's
non-native accent, or the topic and vocabulary. The technology can be used for
the target languages, for which human interpretation is unavailable, e.g. due to
capacity reasons.

Candidate systems for simultaneous speech translation
 differ
in quality of translation, latency and the approach to stability. Some are
streaming, only adding more words \parcite{grissom-ii-etal-2014-dont,
gu-etal-2017-learning,
arivazhagan-etal-2019-monotonic, AttentionArxiv, DuTongChuanArxiv,
ma-etal-2019-stacl, zheng-etal-2019-simpler,iranzo-sanchez-etal-2022-simultaneous}, some allow
re-translation as more input arrives \parcite{muller-etal-2016-lecture, NiehuesNguyenCho2016,
dessloch-etal-2018-kit,
Niehues2018,arivazhagan-etal-2020-translation}. 
Finally, subtitle
presentation options (size of subtitling window, layout, allowed reading time, font
size, etc.) also affect users' impression. 
The combination of the re-translating approach and
limited space for subtitles is challenging because of ``flicker'', i.e. the updates to the text that the user is reading at the moment, has already
read, or that has been scrolled away. The subtitling options
impact the amount of flicker, reading comfort and delay and may affect the general usability. 

\begin{figure}[t]
    \centering
    \includegraphics[width=\linewidth]{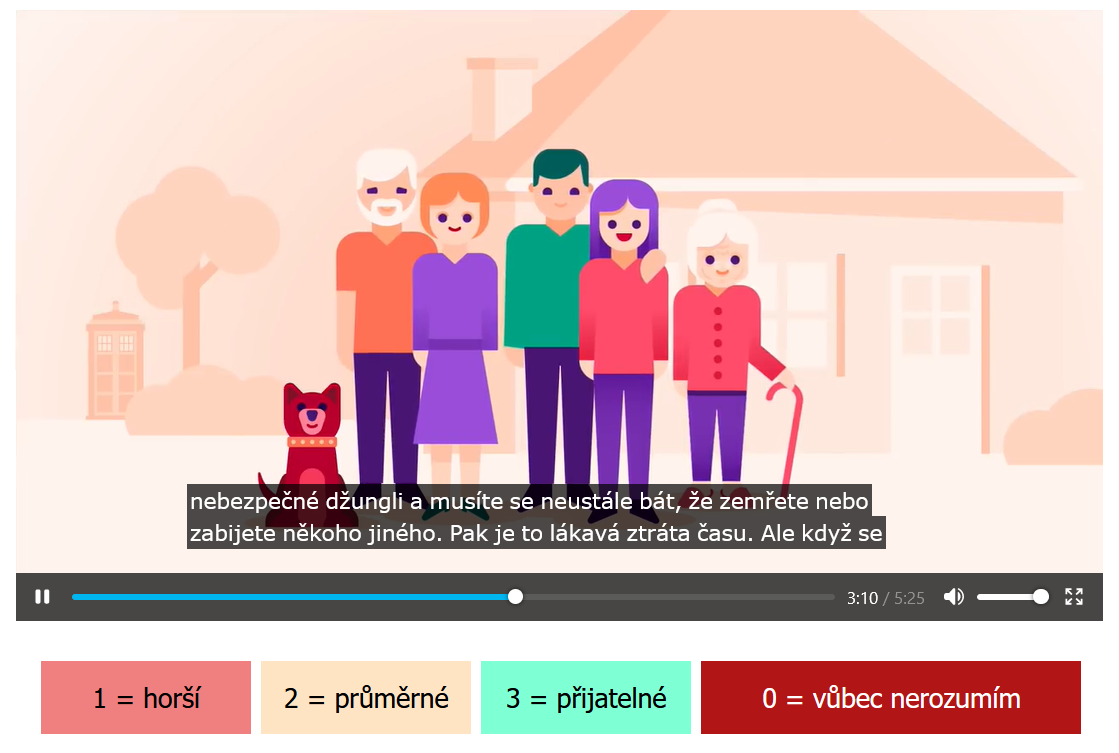}
	\caption{A detail of the default layout with the video document ``Dinge Erklärt: Impfen...''.\footnotemark~The video is at the top, overlaid by two lines of subtitles in Czech, followed by buttons for Continuous Rating. The button labels are: 1: Worse; 2: Average; 3: Good; 0: I do not understand at all.
	}
    \label{fig:exp-screenshot}
\end{figure}

\footnotetext{\url{https://youtu.be/4E0dwFS72gk}}

The evaluation of the traditional, text-to-text machine translation (MT) has been
researched for many years (see e.g. \inparcite{han:mteval-survey:2018} or
developments and discussion
within the series of
WMT, \inparcite{akhbardeh2021findings}). It targets only the translation quality. 
SST evaluation faces new challenges:
simultaneity, latency, and readability to humans.
Evaluating only selected aspects in isolation is reasonable (as MT quality in
\inparcite{elbayad-etal-2020-online}, latency in \inparcite{ma2018stacl, cherry2019thinking}), however, a complete evaluation must be end-to-end, from sound acquisition to subtitling, and take into account the intent of communication. We generalize the intent to passing pieces of information from the speaker (sender) to a participant in an online session (receiver).

\def\contrib#1{(#1)}
\paragraph{Our Contributions}
In this paper, we run an experimental evaluation campaign on 2 hours of documents with German-Czech SST using 32 judges with different levels of source language proficiency. \contrib{i} We contrast two methods of SST evaluation: Continuous Rating and factual questionnaires. We find out that Continuous Rating by bilinguals is easy and reliable for assessing the comprehension.
\contrib{ii} We measure how much comprehension is lost by simultaneity, flicker and presentation options. \contrib{iii} We evaluate different presentation options and layouts and find the most preferred one. \contrib{iv} We find a statistically significant evidence that the users with an advanced, but limited knowledge of the source language reach higher comprehension with low latency subtitles than with large latency and low flicker. \contrib{v} We publish our implementation of the subtitling tool, web application for simulating live events with SST subtitling, and SST human evaluation framework.

Since Continuous Rating is easily applicable to any speech documents, even to those without transcripts and reference translations, and requires minimal time overhead for both preparation and user evaluation, we believe it is suitable to become a standardized way for human manual evaluation of SST.

\section{Related Work}

\citet{hamon-etal-2009-end} propose user evaluation of speech-to-speech
simultaneous translation. To test the adequacy and intelligibility, they
prepared questionnaires with factual questions from the source speech. The
judges listened either to the interpreter, or the machine, and answered the
questions. They evaluated the offline mode, the judges were allowed to stop and
replay the audio while answering. This way the authors measured the
comprehension loss caused by the automatic translation or interpretation. Each
sample was processed by multiple judges, to eliminate human errors. Fluency was
assessed by the judges on a scale.

\citet{machacek2020presenting} propose a technique for collecting continuous user rating
while the user watches video and simultaneous subtitles. The user is asked to
express the satisfaction with the subtitles at any moment by pressing one of
four buttons as the rating changes.

\citet{muller-etal-2016-evaluation} analyzed the feedback from foreign students
using KIT Lecture Translator within two semesters. 
Such a long-term and informal evaluation differs considerably from 
judging in controlled conditions. On one hand, it
summarizes the real-life situation with all the variables and corner cases that
a lab test could only approximate or omit. On the other hand, the users may not
be motivated to give the feedback, and can give only personal opinions that may be
biased. This way it is also difficult to compare multiple system candidates.

\section{Evaluation Campaign}

In our evaluation, we simulate live events on which participants need
assistance with understanding the spoken language. The source and target languages in our study are German and Czech, respectively.
This is an interesting example of two neighbouring countries, distinct language families and yet a relatively well studied pair with sufficient direct training data.

\subsection{Translation System}

We use the ASR system originally prepared for German lectures
\citep{cho-real-world}. It is a hybrid HMM-DNN model emitting partial hypotheses
in real time and correcting them as more context is becoming available. The same system
was used also by KIT Lecture Translator \citep{muller-etal-2016-lecture}.

The system is connected in a cascade with a tool for removing disfluencies and
inserting punctuations \citep{Cho2012SegmentationAP}, and with a German--Czech
NMT system.
\XXXifaccepted{
The cascade is the same as the one of ELITR project at IWSLT 2020
\citep{machacek-etal-2020-elitr}.}

The machine translation is trained on 8M sentence pairs from Europarl and Open
Subtitles \citep{koehn2005europarl,lison-tiedemann-2016-opensubtitles2016}, and validated on newstest.
The Transformer-based \citep{attention-is-all-you-need:2017} system runs in Marian \citep{mariannmt} and reaches 18.8 cased BLEU on WMT newstest-2019.
\XXXifaccepted{
It was described in \citet{kvapilikova-etal-2019-cuni} as the supervised
benchmark.
It is not adapted for stability or for sentence prefixes.
The
translations of unfinished sentences flicker more than the ASR.}

Despite the translations are pre-recorded and only played back in our simulated setup, we ensured we keep the original timing as emitted by the online speech translation system.

\def\Maus{Maus}
\def\DW{DW}
\def\Dinge{Dinge}
\def\Mock{Mock Int}
\begin{table}[t]
    \centering
    \footnotesize
    \begin{tabular}{l|r@{~}r|l}
\bf \parbox{5mm}{Type} & \bf \# & \bf Length & \bf Description \\
\hline
TP&	3&	18:08 &	European Parliament \\
TP&	3&	17:34&	\parbox{45mm}{DG SCIC, Repository for interpretation training} \\[7pt]
A & 3 &	27:52 &	\parbox{45mm}{A mock interpreted conference at interpretation school} \\[7pt]
V& 	2& 	14:43 &	Maus, Educative videos for children \\
A	& 2	& 18:48 & DW, For learners of German \\
V &	2&	16:09 &	Dinge, Educative videos for teens \\
\hline
All & 15 & 114:52 \\
    \end{tabular}
	\caption{Summary of domains of selected documents. Type distinguishes audio only (A), talking person only (TP) and video (V) with
	illustrative or informative content. Length is reported in minutes and seconds.}
    \label{tab:domains}
\end{table}

\subsection{Selection of Documents}

We selected German videos or audio resources that fulfilled the following four conditions:
1) Length 5 to 10 minutes (with some exceptions).
2) The translations had to be of a sufficient quality. Based on a manual check, we discarded several candidate documents: a math lecture and broadcast news
due to many mistranslated
technical terms and named entities. Another group of documents was mistranslated and discarded because they were not long-form speeches, but isolated utterances with long pauses.
3) Informative content. We intend to measure adequacy and comprehension by
asking the judges complementary questions. We thus excluded the documents where the
speaker is not giving information by speech, but uses mostly paralinguistic means, e.g.~singing, poetry, or non-verbal communication.
4) Non-technicality. We expect the judges answer in several plain words in their mother tongue. They
may lack knowledge of any specialized vocabulary.

We selected audios, videos with informative or illustrative content, and
videos of talking persons, to compare user feedback for these types
of documents.
\cref{tab:domains} summarizes the selected documents.

\subsection{Subtitler: Subtitle Presentation}

Subtitler is our implementation of the algorithm by
\citet{machacek2020presenting}
extended by automatic adaptive reading speed in addition to the ``flicker'' parameter as defined in \citet{machacek2020presenting}.
The speed varies between 10
and 25 characters per second 
depending on the current size of the incoming buffer. The default font size is 4.8 mm. The default
subtitling window is 2 lines high and 163 mm wide.\footnote{All typographical properties follow \url{https://bbc.github.io/subtitle-guidelines/}} By default, we use the maximum
flicker and the lowest delay (presenting all translation hypotheses,
not filtering out the partial and possibly unstable ones), no colour
highlighting,
and smooth slide-up animation while scrolling. The example of the setup can be
seen in Figure~\ref{fig:exp-screenshot}.

With the default subtitling window,
90\% of the words in the test documents are finalized in subtitles at most 3 seconds after translation. In 99\%, it is at most 7 seconds. More details and the comparison to fixed reading speed are provided in \cref{sec:appendix_adaptive}.\footnote{The source code of Subtitler is available at \url{https://github.com/ufal/subtitler}}

\subsection{Web Application as Simulation Environment}

We implemented a web application for presenting video and audio documents with embedded Subtitler. We use it for simulation of live subtitled events. The application is equipped with a tool for collecting users' feedback. It also allows administrators to design experiments with different variables (document, subtitling layout, subtitling option) and distribute them to individual judges.\footnote{The source code of the application is available at \url{https://github.com/ufal/continuous-rating}}

\subsection{Types of Feedback}
\label{sec:metrics}

\paragraph{Continuous Rating}
Inspired by \citet{machacek2020presenting}, we add 4 buttons below the audio/video document. While watching, the parti\-ci\-pants are asked to press the buttons to indicate their current satisfaction with the subtitles. We let participants decide the frequency of rating but we suggest clicking each 5-10 seconds or when their assessment has changed. We encourage them to provide feedback as often as possible even if their assessment has not changed. The scores of the rating range between 0 (the worst) and 3 (the best). The order 1, 2, 3, 0 matches the keyboard layout; participants are encouraged to use keyboard shortcuts. The layout is illustrated in \cref{fig:exp-screenshot}.

\paragraph{Questionnaires}
Answering questions as an evaluation approach has been already used \cite{hamon-etal-2009-end, berka2011quiz}. Our questionnaires were composed of two parts: factual questions and general questions.

For \textbf{factual questions} we used the open style, i.e. asking for a short response, instead of yes/no
or multiple choice to exclude guessing. 
We asked a Czech teacher of German to prepare the questions and an answer key
from the original German documents, regardless of the machine translation. The
teacher wrote the questions in Czech, and was instructed to prepare one question
from every 30 seconds of the stream and distribute them evenly, if possible. The
questions had to be answerable only after listening to the document, and not
from the general knowledge. The complexity of the questions was targeted on the
level that an ordinary high-school student could answer after listening to the source document once, if the student would not have any obstacles in understanding
German.
To reduce the effect of limited memory,  the judges had an option in the questionnaire to indicate they knew the answer
but forgot it. Furthermore, they had to fill, from which source they knew the answer: from the subtitles, from the speech, from an image
on the video, or from their previous knowledge.

Finally, we evaluated the factual questions manually against the key, rating them at three levels: correct, incorrect, and partially correct.

After the factual questions, all the questionnaires had a common part with \textbf{general questions} where we
asked the judges on their impression of translation fluency, adequacy,
stability and latency, overall quality, video watching comfort, and a summary comment.

\subsection{Judges}

\begin{table}
    \centering
    \footnotesize
    \begin{tabular}{l|ccccccc|r}
         & \multicolumn{7}{c|}{\bf Layout Experiments} & \\
         \hline
         CEFR & 0 & A1 & A2 & B1 & B2 & C1 & C2 & all \\
         \hline
         Count & 5 & 5 & 1 & 2 & 1 & - & - & 14\\
         \multicolumn{9}{c}{} \\
    \end{tabular}
    
    \begin{tabular}{l|c|cc|cccc|r}
         & \multicolumn{7}{c|}{\bf Flicker Experiments} & \\
         \hline
         & \bf Z & \multicolumn{2}{c|}{\bf Begin.} & \multicolumn{4}{c|}{\bf Advanced} \\
         CEFR & 0 & A1 & A2 & B1 & B2 & C1 & C2 & all \\
         \hline
         Count & 3 & 1 & 3 & - & 2 & 8 & 1 & 18 \\
         \multicolumn{9}{c}{} \\
         \hline
         All & 8 & 6 & 4 & 2 & 3 & 8 & 1 & 32 \\ 
    \end{tabular}
    
    \caption{The judges by their German proficiency levels on CEFR scale and their assignment to experiments. In Flicker experiment, the distribution to groups: Zero level, Beginners, Advanced.}
    \label{tab:judges}
\end{table}

We have conducted two groups of experiments, each with different and distinct groups of judges.

In Comprehension and Layout experiments (\cref{sec:comprehension,sec:layout}), we examined distinct subtitling features. We selected 14 native Czech speakers as judges. Their self-reported knowledge of German had to be between zero and B2 on the CEFR\footnote{Common European Framework of Reference for Languages} scale, to ensure they need some level of assistance with understanding German. We also ensured they do not have knowledge of any other language which could help them understanding German. 

For Flicker experiments (\cref{sec:flicker}), we found other 18 native Czech speakers with an unrestricted German proficiency, to contrast their feedback and level of German. For further analyses, we divided them into three groups.
For brevity further in the paper, we denote the judges with no proficiency of German as ``Zero'' level group, with proficiency between A1 and A2 as ``Beginners'', and the others as ``Advanced''.
See summary of the judges in \Cref{tab:judges}. 

The judges were paid for participation in the study. Each judge spent in total 2 hours on watching and 3 hours on the questionnaires. They watched the videos at their homes on their own devices. They were asked to customize their screen resolution and eye-screen distance to suit their comfort.

\section{Results}
\label{sec:results}

\def\inhandvoting{Offline+voting}
\def\inhand{Offline}
\def\p{$\pm$}
\def\unflicker{Online, without flicker}
\def\best{Online, flicker, top layout}
\def\worst{Online, flicker, least preferred}

First, we analyzed the comprehension levels (\cref{sec:comprehension}) and presentation layouts (\cref{sec:layout}).
Then, we selected the most preferred layout and used it for examining the impact of flicker on comprehension in Flicker experiments (\cref{sec:flicker}).\footnote{The collected data are available at \url{http://hdl.handle.net/11234/1-4913}}

\subsection{Comprehension Levels}
\label{sec:comprehension}

\begin{table}[t]
    \centering
    \footnotesize
    \begin{tabular}{l|ll}
\bf Type & \bf w. avg\p std & \bf $t$-test \\
\hline
\inhandvoting & 0.81\p0.11 & \\
\inhand & 0.59\p0.16 & $^{***}$ \\
\unflicker & 0.36\p0.16 & $^{***}$ \\
\best & 0.33\p0.13 & \\
\worst & 0.31\p0.16 & \\
    \end{tabular}
	\caption{Comprehension scores on all documents and judges. The average weighted by number of questions in document. $^{***}$
	denote the statistically significant
	difference (p-value$<0.01$) between the current and
	previous line.}
    \label{tab:comprehension_wavg}
\end{table}

In our study, we assume comprehension can be assessed as a proportion of correctly
answered questions. We assume the following model: A person without any language
barrier and with non-restricted access to the document during answering the
questionnaire can answer all questions correctly. With a language barrier and
offline MT (unlimited perusal of the document while
answering), some information may be lost in machine translation. More
information is lost with one-shot access to online machine translation because
of forgetting and temporal inattention. Some more information may be lost
because of flicker, and some more because of suboptimal subtitling
layout.

Our results confirm the assumed hierarchy of comprehension levels. Moreover, we
notice that even
the judges with offline MT give inconsistent answers. Combining them and counting answers as correct if at least one judge is correct leads to higher scores. We explain it by insufficient attention. 

\Cref{tab:comprehension_wavg} summarizes the results on all documents. We
measured that on average, 81\% of information was preserved by machine
translation (\inhandvoting, i.e. one of two judges answered correctly). A single judge could find 59\% of information
(\inhand). 
In an oracle experiment without flicker, when the machine translation gives the
final hypotheses with the timing of the partial ones (i.e. as if it knew the
best translation of the upcoming sentence), a single judge could
answer 36\%. In real setup with flicker and the most preferred subtitling layout
(\best), 33\% information was found, and 31\% with less preferred. The standard
deviation is between 11 and 16\%.

We found statistically significant difference (two-sided $t$-test) between offline
MT with voting and without it, and between offline and online MT.

\subsection{Layout Preference}
\label{sec:layout}

We analyzed effects of distinct subtitling features by contrastive experiments
differing only at one feature, see the paragraphs in this section. We distributed them randomly among the judges,
regardless of their German skills. After watching each document, the judge fills the questionnaire.

In all cases, the results show a slight insignificant preference towards one
variant of the feature in all three types of feedback that we collect: ``Comprehension'' is the proportion of correctly answered factual questions, ``Averaged Continuous Rating'' is an averaged feedback from button clicks, and ``Final rating'' summarizes the responses in the general section of questionnaires.

For visually informative videos, we separately report the scores of ``Watching  comfort'' which we collected in the general section of questionnaires. Some judges provided also textual feedback, examples are in \cref{sec:appendix_textual_feedback}.

\def\side{Side}
\def\shortmiddle{Below}
\def\longmiddle{Long, middle}
\def\all{sum, avg}
\def\Final{\parbox{16mm}{\centering Final rating}}
\def\Comprehension{\parbox{16mm}{\centering Compre\-hension}}
\def\AvgCont{\parbox{16mm}{\centering Avg. Cont. Rating}}
\def\Watching{\parbox{16mm}{\centering Watching comfort}}
\begin{table*}[t]
    \centering
    \tablesize
\begin{tabular}{l@{~}l||r@{~}r|r@{~}r||r@{~}r|r@{~}r||r@{~}r|r@{~}r}
& &  \multicolumn{4}{c||}{Side vs Below} &  \multicolumn{4}{c||}{Below vs Overlay} &   \multicolumn{4}{c}{Size of subtitling window}  \\
 & &  \multicolumn{2}{|c|}{\side} & \multicolumn{2}{|c||}{\shortmiddle} & \multicolumn{2}{|c|}{Below} & \multicolumn{2}{|c||}{Overlay} & \multicolumn{2}{|c|}{2 l.$\times$163mm} & \multicolumn{2}{|c}{5 l.$\times$200mm}\\
 \hline
\multirow{4}{*}{\Final}
& audio    & & & & & & & & & 10 & 1.80 \p 0.87 & 8 & \bf 2.75 \p 0.97\\  
& talking  &  5 & \bf 2.80 \p 1.33  &  7 & 2.43 \p 1.05 & 9 & 2.33 \p 1.05 & 9 & \bf 2.78 \p 1.13 & 9 & 2.33 \p 1.05 & 5 & \bf 2.80 \p 1.60 \\  
& video    &  1  &  1.00  \p  0.00  &  3 & \bf 1.67 \p 0.94 & 5 & 1.40 \p 0.80 & 8 & \bf 2.38 \p 0.86 & 5 & 1.40 \p 0.80 & 3 & \bf 2.33 \p 0.47 \\  
& \all     &  6 & \bf 2.50 \p 1.38  &  10 & 2.20 \p 1.08 & 14 & 2.00 \p 1.07 & 17 & \bf 2.59 \p 1.03 & 24 & 1.92 \p 1.00 & 16 & \bf 2.69 \p 1.16 \\  
\hline
\multirow{4}{*}{\Comprehension}
& audio    & & & & & & & & & 10 & 0.25 \p 0.15 & 8 & \bf 0.31 \p 0.15 \\  
& talking  &  5 & \bf 0.34 \p 0.25  &  7 & 0.28 \p 0.27 & 9 & 0.29 \p 0.25 & 9 & \bf 0.39 \p 0.20 & 9 & 0.29 \p 0.25 & 5 & \bf 0.40 \p 0.21 \\  
& video    &  1  &  0.18  \p  0.00  &  3 & \bf 0.36 \p 0.04 & 5 & 0.26 \p 0.14 & 8 & \bf 0.37 \p 0.11 & 5 & 0.26 \p 0.14 & 3 & \bf 0.28 \p 0.05 \\           
& \all     &  6 & \bf 0.31 \p 0.24  &  10 & 0.30 \p 0.23 & 14 & 0.28 \p 0.21 & 17 & \bf 0.38 \p 0.17 & 24 & 0.26 \p 0.19 & 16 & \bf 0.33 \p 0.16 \\  
\hline
\multirow{4}{*}{\AvgCont}
& audio    & & & & & & & & & 10 & 0.90 \p 0.71 & 8 & \bf 1.66 \p 0.95 \\  
& talking  &  5 & 1.56 \p 1.00  &  7 & \bf 1.78 \p 0.35 & 9 & 1.65 \p 0.52 & 9 & 1.65 \p 0.99 & 9 & \bf 1.65 \p 0.52 & 5 & 1.09 \p 0.78 \\  
& video    &  1  &  0.23  \p  0.00  &  3 & \bf 1.21 \p 0.45 & 5 & 1.11 \p 0.50 & 8 & \bf 1.15 \p 0.77 & 5 & 1.11 \p 0.50 & 3 & \bf 1.35 \p 0.31 \\  
& \all     &  6 & 1.33 \p 1.04  &  10 & \bf 1.64 \p 0.45 & 14 & \bf 1.47 \p 0.57 & 17 & 1.42 \p 0.93 & 22 & 1.21 \p 0.70 & 16 & \bf 1.42 \p 0.85 \\  
\hline
\multirow{3}{*}{\Watching}
& talking  &  5 & 2.80 \p 0.75  &  7 & \bf 3.33 \p 0.75 & 9 & 3.43 \p 0.73 & 9 & \bf 4.11 \p 0.74 & 7 & \bf 3.43 \p 0.73 & 5 & 2.80 \p 0.98 \\  
& video    &  1  &  2.00  \p  0.00  & 3 & \bf 3.00 \p 1.63 & 5 & 2.20 \p 1.60 & 8 & \bf 3.00 \p 1.00 & 5 & 2.20 \p 1.60 & 3 & \bf 2.33 \p 1.25 \\  
& \all     &  6 & 2.67 \p 0.75  & 10 & \bf 3.22 \p 1.13 & 14 & 2.92 \p 1.32 & 17 & \bf 3.59 \p 1.03 & 12 & \bf 2.92 \p 1.32 & 8 & 2.62 \p 1.11 \\  
\end{tabular}
	\caption{Results of the contrastive experiments for Side vs Below, Below vs Overlay and Subtitling window size: 2 lines height $\times$ 163 mm width vs 5 lines height $\times$ 200 mm width.  The three numbers in each row and cell are
	the number of experiments, average and standard deviation. The higher score,
	the better. Comprehension rate is between 0 and 1, average continuous rating
	is between 0 and 3, the others on a discrete scale 1 to 5. Higher score in each experiment is bolded. The last row of each section summarizes the scores across document types.}
    \label{tab:side-below}
\end{table*}

\begin{description}[style=unboxed,leftmargin=0cm]
\item[Side vs Below]~~For videos and videos with a talking person, we consider two locations for the subtitle window:
on the left side of the video,
or below. The side window can be high but narrow (17 lines of 60 mm width, to match the
height of the video), while the window underneath is short and wide (2 lines of 163
mm width). The first is more comfortable for reading, the latter for watching the video.
    
The results are in \Cref{tab:side-below} on the left. There is a preference for the layout ``below'' when the video is informative, and for ``side" otherwise.

\item[Below vs Overlay]~~The subtitling window can be placed over the video, as in films, or below. In
the first case, the subtitles possibly hide an informative image content, in
the latter case, there is a larger distance between the image and the subtitles.
The results on non-German speaking judges are insignificantly in favor of
overlay, see the middle of \cref{tab:side-below}.

\def\Size{Size~[lines,mm width]}

\def\Short{2$\times$163}
\def\Middle{5$\times$200}
\def\Long{18$\times$250}

\begin{table*}[t]
    \centering
    \tablesize
\def\yes{Yes}
\def\no{No}
\begin{tabular}{l||r@{~}r|r@{~}r||r@{~}r|r@{~}r||r@{~}r|r@{~}r}
Highlighting &  \multicolumn{2}{c|}{\no} &  \multicolumn{2}{c||}{\yes}  &  \multicolumn{2}{c|}{\no} &  \multicolumn{2}{c||}{\yes}  & \multicolumn{2}{c}{\no} & \multicolumn{2}{c}{\no} \\
\Size  &  \multicolumn{4}{c||}{18$\times$250 (``Large'')} &  \multicolumn{4}{c||}{5$\times$200 (``Medium'')} &   \multicolumn{2}{c|}{\Long} &  \multicolumn{2}{c}{\Middle}  \\
\hline
\multirow{1}{*}{Final~rating}       &  14  &  2.93  \p  0.80  &  13  & \bf 3.31  \p  1.14 &    2  &  2.50  \p  0.50  &  1  & \bf 4.00  \p  0.00  & 11 & \bf 2.91 \p 0.79 & 8 & 2.75 \p 0.97  \\
\multirow{1}{*}{Comprehension}      &  14  &  0.25  \p  0.15  &  13  & \bf 0.30  \p  0.12 & 2  &  \bf 0.44  \p  0.18  &  1  &  0.39  \p  0.00 & 11 & 0.23 \p 0.14 & 8 & \bf 0.31 \p 0.15   \\
\multirow{1}{*}{Avg.~Cont.~Rating}  &  14  &  1.32  \p  0.82  &  13  & \bf 1.42  \p  0.74 &    2  & \bf 2.19  \p  0.50  &  1  &  2.12  \p  0.00 & 11 & 1.50 \p 0.79 & 8 & \bf 1.66 \p 0.95   \\
\multicolumn{5}{c}{~} \\
\end{tabular}
    \caption{Results of highlighting experiments on audio documents and subtitling window size 5 lines $\times$ 200 mm vs 18 lines $\times$ 250 mm. Description of numbers as in \Cref{tab:side-below}.}
    \label{tab:colours}
\end{table*}

\item[Highlighting Flicker Status]~~The underlying rewriting speech translation system distinguishes three levels of status for segments (automatically identified sentences):
``Finalized'' segments no longer change.
``Completed'' segments are sentences which received
a punctuation mark. They can be changed by a new update and the prediction of the punctuation may also change or disappear.
They usually flicker once in several seconds.
``Expected'' segments are incomplete sentences, to which new translated
words are still appended. They flicker several times per second.

It is a user interface question if the status of the segments should be indicated by highlighting, or if this piece of information would be rather disturbing. We experimented only with colouring text background in large and medium subtitling window for audio-only documents.
    
Our experiments show that the judges prefer highlighting flicker status in the large window. For the medium window, this inclination is less clear, see \cref{tab:colours}.

\item[Size of Subtitling Window]~~The subtitling window can be of any size. If the window is short and narrow,
there is a short gap between an image and subtitles, which simplifies focus
switching. On the other hand, a small window contains short history, so the
user can miss translation content if it disappears while paying attention to
the video. A small window may also accidentally cause a long subtitling delay if the 
translation was updated in the scrolled-away part of text. In this situation, Subtitler has to ``reset'' the subtitles and repeat the part.
With a large window, the distance between the growing end of the subtitles and the image is larger. The content stays longer, but it is more complicated to find
a place where the user stopped reading before the last focus switch.

Depending on spatial constraints, it is always recommended to use as large window as possible,
especially for documents without visual information, where focus switching
between an image and subtitles is not expected. 
We tested two pairs of sizes on the same documents. The results are in \Cref{tab:side-below} on the right. As
we expected, the window with 5 lines was rated insignificantly better than with 2 lines in most scales and setups, but the 2-line reached a higher average watching comfort (2.92) that the 5-line setup (2.62).

For an audio-only document, we also tested the large (18 lines) vs. medium (5 lines) window, observing users' reported preference for the large one but slightly higher comprehension and continuous feedback for the medium one, see the right part of \cref{tab:side-below}.

\end{description}

\subsection{Flicker Experiments}
\label{sec:flicker}

We assume that the user behaviour differs by knowledge of the source language.
We hypothesize that the Zero group of users and Beginners read all the subtitles all the time and do not pay
attention to the speech. They do not mind large latency, 
but demand high quality translation, and comfortable reading
without flicker.
On the other hand, the users with an advanced knowledge of the source language may listen to the speech, try to understand on their own, and
look at the subtitles only occasionally, when they are temporarily uncertain or
need assistance with an unfamiliar word. They need low latency, and do not mind
slightly lower quality.

To empirically test our hypothesis, we prepared two realistic setups: With flicker, the subtitles
are presented immediately as available, but with frequent rewriting which
discomforts the reader. Without flicker, the translations are delayed until the SST system confirms they will not change, and that usually happens during uttering the next sentence. We selected two
videos for this experiment and distributed these setups uniformly between all groups of judges.

The results of comprehension are in \Cref{tab:german}. It shows that
Advanced users achieve higher comprehension with flicker (58\%) than without (49\%). We found the difference statistically significant, which confirms the second part of our hypothesis.

The Zero level speakers and Beginners also report higher comprehension with flicker
(Zero: 30\% vs 34\% and Beginners: 31\% vs 33\%), but this difference is statistically insignificant. Even though the preference inclines towards flicker, it is less noticeable compared to the Advanced group, and we consider this difference negligible. The other types of feedback (Average Continuous Rating and Overall rating from the end of questionnaire; not shown) confirm the trend of Comprehension for all groups.

\begin{table}[t]
\centering
\footnotesize
\begin{tabular}{l|r@{~}r|r@{~}r|r@{~}r}
                         & \multicolumn{2}{c|}{\bf Zero level}      & \multicolumn{2}{c|}{\bf Beginners} & \multicolumn{2}{c}{\bf Advanced} \\\hline
\multicolumn{1}{l|}{flic.} & 27 & \bf 0.34 \p 0.16 & 33 & \bf 0.33 \p 0.16 & 91 & \bf 0.58 \p 0.19 \\ 
\multicolumn{1}{l|}{no f.}  & 29 & 0.30 \p 0.15 & 38 & 0.31 \p 0.12 & 81 & 0.49 \p 0.20\\
\hline
\multicolumn{1}{l|}{}      & \multicolumn{2}{c|}{insignificant} & \multicolumn{2}{c|}{insignificant} & \multicolumn{2}{c}{$p<0.01$} \\
\end{tabular}
	\caption{Comprehension scores on a setup with flicker and no flicker, as rated by judges with different source language proficiency. The three numbers in each row and cell are the number of samples, average and standard deviation. Higher scores bolded. The difference between setups within Advanced group is statistically significant with $p<0.01$.
}
    \label{tab:german}
\end{table}

\subsection{Comprehension vs Continuous Rating}
\label{sec:relation}

\begin{figure*}
    \centering
    \includegraphics[width=\linewidth]{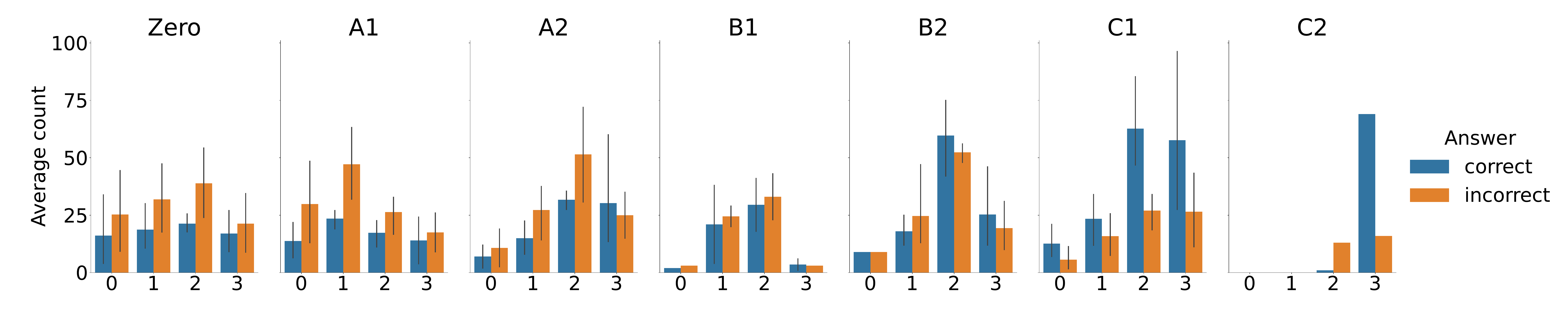}
    \includegraphics[width=\linewidth]{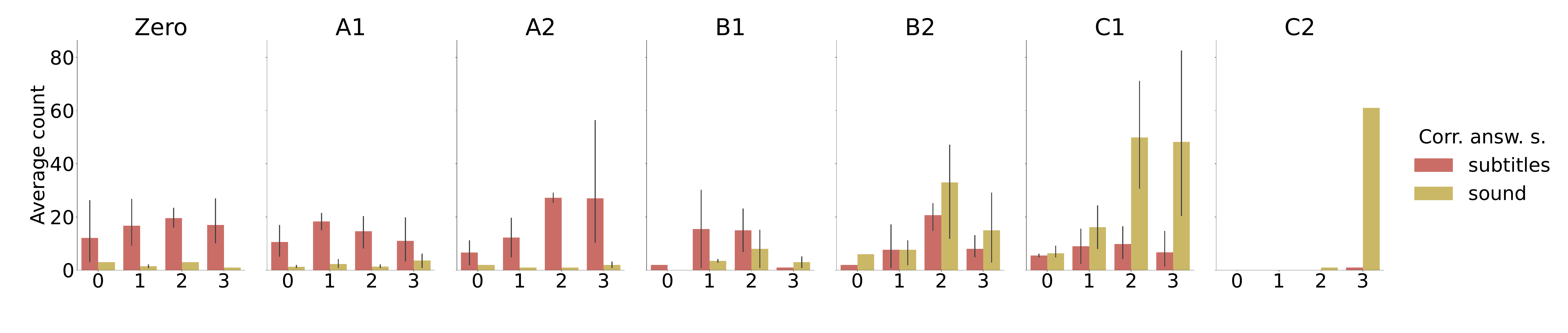}
    \caption{
    The average count of answers per judge for each proficiency level. Top: Correct (OK/OK-, blue bars) and incorrect (wrong/unknown, orange bars) answers vs Continuous Rating at the time when the answer was disclosed in the original document (x-axis, 0 means worst, 3 the best),
    distributed by source language proficiency level of the judges.
    Bottom: From which source the judges learned the correct or partially correct answer; subtitles in red, sound in yellow.
    }
    \label{fig:cont-answers}
\end{figure*}

\begin{table}
    \centering
    \footnotesize
    \begin{tabular}{l|ccc}
& \multicolumn{3}{c}{\bf $\chi^2$-test $p$-values}  \\
 & \bf Zero level & \bf Beginners & \bf Advanced \\
\hline
OK/OK- & 0.24& $\bf 1.8\cdot10^{-5}$ & $\bf 5.6\cdot10^{-5}$ \\
unknown & 0.033& $\bf 1.7\cdot10^{-4}$ & $\bf 9.1\cdot10^{-4}$ \\
wrong & 0.59 & 0.45 & $\bf 2.9\cdot10^{-3}$ \\
forgot & 0.9 & 0.48 & 0.019 \\
    \end{tabular}
    \caption{The results of $\chi^2$-test for independence of Continuous Rating and answer correctness. Bolded values are where the two variables are \textbf{dependent} with statistical significance $p < 0.01$.
    }
    \label{tab:chi_squared}
\end{table}

We collected Continuous Rating of the overall quality of subtitles at given
times.
For every comprehension question, we know the time span when the answer appears in the source speech document.
Based on this timing information, we can relate comprehension and Continuous Rating.
For a given time span answering a particular question, we find the most frequent Continuous Rating (button clicked most often) for every annotator. This gives us a histogram of Continuous Rating scores reported by different judges.
In \cref{fig:cont-answers} top, we show the correct (``OK'') or partially correct answers (``OK-'') and the 
histogram of Continuous Ratings by judges of distinct German proficiency levels. For a more detailed plot including all evaluation classes see \cref{sec:appendix_detail}. This data aggregates observations
for all documents and all setups excluding
the offline SST and the oracle online SST without flicker.

For the judges with zero knowledge of German, we can not see any dependency of their comprehension to their Continuous Rating.
On the other hand, the more the judges are proficient in German, the more their Continuous Rating reflects their comprehension. For example, for the C1 judges (Advanced) we can estimate their comprehension (and thus subtitle quality) from their clicking well: When they understand the content, the most probable given rating is 3 or 2. A less probable rating is 1, and they almost never rate 0 when they understand the content. 

\paragraph{Listening while Rating}
In \cref{fig:cont-answers} bottom, we show, from which source the judges knew the correct answer, either from the subtitles, or from sound. We can observe that indeed, the judges with German proficiency level B1 and higher listen to the source sound and understand, while the Zero level judges and Beginners rely only on subtitles.

\paragraph{Statistical Test}
To test the relation rigorously, we divide the judges into three groups by proficiency levels, their counts (see \cref{tab:judges}), their relation of Continuous Rating to correct answers and approach to listening versus reading (\cref{fig:cont-answers}).
We run $\chi^2$-test for statistical independence of Continuous Rating and answer results on the three groups. Test results are in \Cref{tab:chi_squared}. It shows that for the judges with Zero level of German, their Continuous Rating is independent on answer results. They
do not follow the sound at all because they do not understand it, and rate only the readability and flicker. In case of the Beginners (A1 and A2, recall \cref{tab:judges}), we observe the dependency of their Continuous Rating on correct answers (``OK/OK-'') and on cases when they did not answer (``unknown''). Their wrong answers and forgetting is independent of Continuous Rating, they probably make random mistakes uniformly.
The Advanced group of judges give their correct, unknown or wrong answers consistently with their Continuous Rating. We therefore assume that they follow and understand the source speech and include the adequacy in their Continuous Rating.

We can also see that in all the three groups, the forgotten answers are independent on Continuous Rating. We assume that random and uniform outages may be characteristic for human memory.

\paragraph{Practical Conclusions}
We conclude that Continuous Rating is a suitable for manual evaluation of simultaneous machine translation.
The judges who speak the source
language on at least B2 level on CEFR scale have an ability to assess SST quality reliably only by Continuous Rating, without the need for questionnaires which are laborious to prepare, answer and evaluate. 

\section{Conclusion}

We proposed a novel and effective method for end-to-end user evaluation of simultaneous speech
translation SST called Continuous Rating, publishing an open source evaluation tool for the future use. We showed that this method can be used for measuring comprehension and evaluating subtitling parameters. We demonstrated how user comprehension differs from offline MT to online MT. We showed that the users with a knowledge of the source language prefer low latency despite higher instability. We demonstrated that Continuous Rating can be used as a time-efficient human evaluation metric when employing judges with at least B2 (or, preferrably, C1) level of source language proficiency.

\section*{Limitations}

This work is limited to only one direction of SST and lacks the comparison of multiple SST variants. Additionally, due to the number of investigated subtitling features and the smaller sample of judges, the results of layout experiments show only statistically insignificant preference towards one variant.

\section*{Acknowledgments}

The research was partially supported by the grants 19-26934X  (NEUREM3)  of  the  Czech  Science Foundation, 
``Grant Schemes at CU'' (reg. no. CZ.02.2.69/0.0/0.0/19\_073/0016935) and
SVV project number 260~575.

\bibliographystyle{acl_natbib}
\bibliography{anthology,acl2021,other}

\appendix

\section{Subtitler}

\subsection{Adaptive Reading Speed: Delay}
\label{sec:appendix_adaptive}

\begin{table}[ht]
    \setlength{\tabcolsep}{4.5pt}
    \centering
    \footnotesize
    \begin{tabular}{l|ccccc|c|c}
        & & \multicolumn{4}{c}{Delay} & & \\
        & 70\% & 80\% & 90\% & 95\% & 99\% & max & resets \\\hline
        ARS & \bf 0.01 & \bf 1.44 & \bf 3.06 & \bf 4.51 & \bf 7.05 & \bf 12.06 & 8.80 \\
        FRS & 1.74 & 3.54 & 5.18 & 7.52 & 10.65  & 16.78 & \bf 5.47 \\
    \end{tabular}
    \caption{The adaptive reading speed (ARS) in comparison to the fixed reading speed (FRS), set to 18 char/sec. Percentages denote the proportion of words that have a delay less than the given number. The delay is in seconds, resets in the average count per document.}
    \label{tab:adaptive_speed}
\end{table}

We compared adaptive to fixed reading speed, averaging over all documents.
We set the value of fixed reading speed to 18 characters per seconds, which we obtained by averaging all delays in the setting without adaptive reading speed.

The comparison is in \cref{tab:adaptive_speed}. The delay was measured for all presented words. We used a subtitling window of 2 lines $\times$163 mm because it represents an upper bound for the delay of bigger subtitling windows.

\section{Results}

\subsection{Comprehensions vs Continuous Rating}
\label{sec:appendix_detail}

\begin{figure*}[t]
    \centering
    \includegraphics[width=0.22\linewidth]{CEFR-counts-answers-vs-rating-big-noy.pdf}
    \includegraphics[width=0.22\linewidth]{CEFR-OK-counts-source-vs-rating-big-noy.pdf}
    \includegraphics[width=0.22\linewidth]{CEFR-all-counts-source-vs-rating-big-noy.pdf}
    \caption{
    The average count of answers per judge for each proficiency level. Left: OK, OK-, wrong, unknown and forgotten answers vs Continuous Rating at the time when the answer was disclosed in the original document (x-axis, 0 means worst, 3 the best), distributed by source language proficiency level of the judges: from zero through beginners (A1, A2) and intermediate (B1, B2) to advanced (C1, C2). Middle: From which source the judges learned the correct (OK) or partially correct (OK-) answer. Right: From which source the judges learned all answers, regardless of their evaluation.
    %\XXX{přepsat podle Fig 2} For each level of CEFR scale, extended plots for the distribution of Continuous Rating and results of answers (left), the distribution of Continuous Rating and source of all answers (middle) and the distribution of Continuous Rating and source of OK/OK- answers (right). \XXX{předělat jako u Fig 2. Zprůměrovat ty judge do jednoho. Popisky x-ové osy u všech. \repl{None}{Zero}}
    }
    \label{fig:cont-answers-detail}
\end{figure*}

In \cref{fig:cont-answers-detail}, we show the average count of answers per judge for each proficiency level.
Note two observations: 1) The number of already known answers is negligible, which proves that the questions were selected based on the content of documents. 2) The number of answers whose source was not given is high for all answers (\cref{fig:cont-answers-detail}, right column), whereas it is low when correct and partially correct answers were selected (\cref{fig:cont-answers-detail}, middle column). It means that judges provided the source when they answered a question.

\subsection{Textual Feedback}
\label{sec:appendix_textual_feedback}

\begin{table*}[t]
\centering
\tablesize
\begin{tabular}{p{0.07\textwidth}|p{0.8\textwidth}}
 & \multicolumn{1}{c}{\bf Feedback} \\\hline\hline
\bf Setting & \multicolumn{1}{c}{\bf C1 proficiency, Overlay layout}\\\hline
\multirow{5}{*}{Flicker} & The subtitles weren't so bad in terms of content or latency. % Titulky nebyly z hlediska obsahu ani časového posunu tak špatné
\\\cline{2-2}
& The subtitles were very good, they just got stuck in the middle of the video, but after a short pause they worked again without any problems. % titulky byly moc dobré, akorát se někdy v půlce videa zasekly, ale po krátké pauze zase fungovaly bez problému
\\\cline{2-2}
& The subtitles were relatively good, but despite their intelligibility and relative linguistic accuracy, they seemed very chaotic and very uncomfortable to read. % Titulky byly poměrně dobré, ale i přes srozumitelnost a relativní jazykovou správnost působily velmi chaoticky a jejich čtení bylo velmi nepohodlné.
\\\hline
\multirow{5}{*}{No flicker} & A big delay of subtitles was sometimes inconvenient. If the subtitles are very delayed, it is almost impossible to follow them. % Nepohodlné bylo místy velké zpoždění titulků. Pokud jsou titulky hodně zpožděné, nejde se na ně soustředit už skoro vůbec.
\\\cline{2-2}
& The subtitles were small and dense, it was hard to orientate, especially when they were even delayed. %titulky byly malé a nahuštěné, je těžké se v nich zorientovat, zvlášť ve chvíli, kdy jsou ještě navíc zpožděné
\\\cline{2-2}
%& The delay is too large, one has to do 3 things at the same time; listening to the source, reading the translation and evaluating the translation. Then, one hardly catches the message of presentation. % zpoždění je příliš veliké, člověk musí dělat 3 věci zároveň, poslouchat originál, číst překlad a hodnotit překlad, poté utíká člověku význam sdělení
%\\\cline{2-2}
& At first, the delay was small. Then, at one point the subtitles got stuck and there was a lot of delay behind the sound. % Ze začátku bylo zpoždění malé, pak se v jednom bodě titulky zasekly a hodně se zpožďovaly za zvukem 
\\
\end{tabular}
\caption{The selection of textual feedback from judges.}
\label{tab:feedback}
\end{table*}

In \cref{tab:feedback}, we depict several textual ratings from Flicker Experiment. We select judges with C1 source language proficiency and contrast their feedback for flicker and no flicker.

The judges report higher satisfaction with flicker. They notice increased latency when the presentation mitigate flicker. This is consistent with our findings in Flicker experiment for Advanced group.

\end{document}